\def\year{2021}\relax
\documentclass[letterpaper]{article} 
\usepackage{aaai21}  
\usepackage{times}  
\usepackage{helvet} 
\usepackage{courier}  
\usepackage[hyphens]{url}  
\usepackage{graphicx} 
\usepackage{natbib}  
\usepackage{caption} 
\usepackage[switch]{lineno} 
\usepackage{multirow}
\usepackage{amssymb}
\usepackage{amsmath}
\usepackage{bm}

\usepackage{cleveref}

\newcommand{\ie}{\emph{i.e.}}
\newcommand{\eg}{\emph{e.g.}}

\title{Boundary Proposal Network for Two-Stage Natural Language Video Localization}
\author {
        Shaoning Xiao,\textsuperscript{\rm 1}
        Long Chen,\textsuperscript{\rm 2}\thanks{Long Chen is the corresponding author.}
        Songyang Zhang,\textsuperscript{\rm 3}
        Wei Ji,\textsuperscript{\rm 4}
        Jian Shao,\textsuperscript{\rm 1}
        Lu Ye,\textsuperscript{\rm 5}
        Jun Xiao\textsuperscript{\rm 1} \\
}
\affiliations {
    \textsuperscript{\rm 1} DCD Lab, College of Computer Science, Zhejiang University \\
    \textsuperscript{\rm 2} Tencent AI Lab, Shenzhen \\
    \textsuperscript{\rm 3} University of Rochester  \\
    \textsuperscript{\rm 4} National University of Singapore \\
    \textsuperscript{\rm 5} Zhejiang University of Science and Technology \\
    \{shaoningx, jiwei, jshao, junx\}@zju.edu.cn, zjuchenlong@gmail.com, szhang83@ur.rochester.edu, yelue@zust.edu.cn
}

\begin{document}
\maketitle

\begin{abstract}
We aim to address the problem of Natural Language Video Localization (NLVL) --- localizing the video segment corresponding to 
a natural language description in a long and untrimmed video. 
State-of-the-art NLVL methods are almost in one-stage fashion, which can be typically grouped into two categories:
1) anchor-based approach: it first pre-defines a series of video segment candidates (\eg, by sliding window), and then does classification for each candidate; 
2) anchor-free approach: it directly predicts the probabilities for each video frame\footnote{In this paper, the frame is a general concept for an actual video frame or a video clip which consists of a few consecutive frames.}
as a boundary or intermediate frame inside the positive segment. 
However, both kinds of one-stage approaches have inherent drawbacks: the anchor-based approach is susceptible to the heuristic rules, further limiting the capability of handling videos with variant length.
While the anchor-free approach fails to exploit the segment-level interaction thus achieving inferior results.
In this paper, we propose a novel \emph{Boundary Proposal Network (BPNet)}, a universal two-stage framework that gets rid of the issues mentioned above. 
Specifically, in the first stage, BPNet utilizes an anchor-free model to generate a group of high-quality candidate video segments with their boundaries.
In the second stage, a visual-language fusion layer is proposed to jointly model the multi-modal interaction between the candidate and the language query, followed by a matching score rating layer that outputs the alignment score for each candidate. 
We evaluate our BPNet on three challenging NLVL benchmarks (\ie, Charades-STA, TACoS and ActivityNet-Captions). 
Extensive experiments and ablative studies on these datasets demonstrate that the BPNet outperforms the state-of-the-art methods.
\end{abstract}

\section{Introduction}

\begin{figure}[t]
    \centering
    \includegraphics[width=0.475\textwidth]{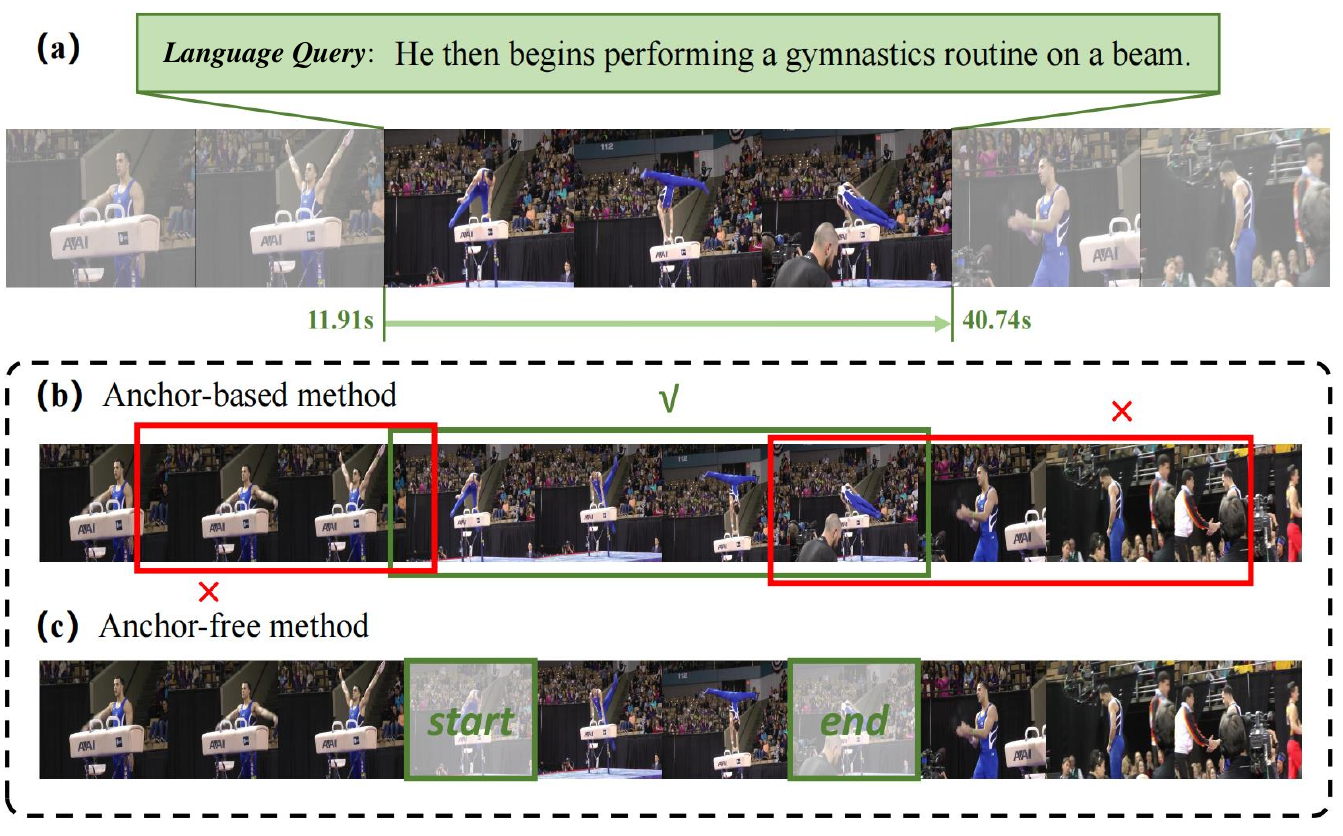} 
    \caption{(a) An illustrative example of the NLVL task.
    Given a video and a query, NLVL is to localize the video segment corresponding to the query with the 
    start point (11.91s) and the end point (40.74s).
    (b) Anchor-based approach: A number of temporal bounding boxes are placed on the video as candidates 
    and the best-matching one (\eg, the green one) is chosen as the result. 
    (c) Anchor-free approach: Each frame is determined whether it is the boundary frame.
    }
    \label{figure1}
\end{figure}

Understanding video content with the aid of natural language, \eg, describing the video content by natural language or 
grounding language in the video, has drawn considerable interest in both computer vision and natural language processing communities. 
This kind of task is challenging since it needs to not only understand the video and the sentence separately but also their corresponding interaction.
Recently, a core task of this area called \textbf{Natural Language Video Localization (NLVL)}~\cite{DBLP:conf/iccv/GaoSYN17,DBLP:conf/iccv/HendricksWSSDR17} has been proposed. 
As shown in Figure~\ref{figure1} (a), given an untrimmed video and a natural language query, NLVL aims to localize the video segment relevant to the query by determining the start point and the end point. 
NLVL is a worth exploring task due to its potential applications, \eg, video content retrieval~\cite{DBLP:conf/eccv/ShaoXZHQL18} and video question answering~\cite{DBLP:conf/emnlp/LeiYBB18,xiao2020hierarchical,ye2017video}.

A straightforward solution for NLVL is the \emph{anchor-based approach}~\cite{DBLP:conf/iccv/GaoSYN17,DBLP:conf/emnlp/HendricksWSSDR18,DBLP:conf/mm/LiuWN0CC18,DBLP:conf/aaai/ChenJ19a,DBLP:conf/wacv/GeGCN19,DBLP:conf/aaai/Xu0PSSS19,DBLP:conf/cvpr/ZhangDWWD19,ChenCMJC18,DBLP:conf/aaai/Wang0J20},
which follows the same spirit of anchor-based object detectors, \eg, Faster R-CNN~\cite{DBLP:conf/nips/RenHGS15}.
Specifically, this kind of method places a number of size-fixed bounding boxes on the temporal dimension of the video,   
and then matches each candidate with the sentence in a common feature space, as shown in Figure~\ref{figure1} (b).
They model the segment-level information of proposals and transforms the localization problem into a multi-modal matching problem.
However, it is worth noting that they suffer from two inherent drawbacks:
(1) In order to achieve higher recall, a vast number of proposals are required, which makes the subsequent matching process inefficient. 
(2) Moreover, they have to elaborately design a series of hyper-parameters (\eg, the temporal scales and sample rate) of the bounding boxes so that they can be adaptable to video segments with arbitrary length.

To tackle these issues, another type of solution called the  \emph{anchor-free approach}~\cite{DBLP:conf/aaai/ChenLTXZTL20,DBLP:conf/aaai/ChenJ19a,DBLP:conf/aaai/YuanM019,LuCTLX19,DBLP:conf/acl/ZhangSJZ20,DBLP:conf/wacv/OpazoMSLG20,DBLP:conf/cvpr/MunCH20} has been proposed. 
As shown in Figure~\ref{figure1}(c), instead of depicting the probable video segments as temporal bounding boxes, anchor-free methods directly predict the start and the end boundaries of the query-related video segment or predict the positive frames between the ground-truth boundaries. 
Benefit from this design, 
the anchor-free methods get rid of placing superfluous temporal anchors, \ie, they are more computation-efficient. 
They are also flexible to adapt to diverse video segments without the assumption of the position and the length of the ground-truth video segment.
Unfortunately, despite these advantages, there is one main factor that strictly limits the performance of anchor-free methods: they overlook the rich information between start and end boundaries 
because they are hard to model the segment-level interaction.

In this paper, we propose a two-stage end-to-end framework termed \emph{Boundary Proposal Network (BPNet)}, 
which inherits the merits of both the anchor-based and anchor-free methods and avoids their defects. 
Specifically, we first generate several high-quality segment proposals using an anchor-free backbone to avoid redundant candidates, then an individual classifier is proposed to match the proposals with the sentence by predicting the matching score.
Compared to anchor-based methods with abundant handcrafted proposals, our design decreases the number of candidates thus alleviating the redundant computation burden.
By utilizing an anchor-free method to generate proposals, our approach can be adaptable to video segments of arbitrary lengths without designing the heuristic rules.
In addition, compared with anchor-free methods, our BPNet is capable of better modeling the segment-level information via a visual-language fusion module.
Furthermore, our proposed framework is a universal paradigm, which means each stage of the framework can be replaced by any stronger anchor-free and anchor-based models to further boost the performance.

We demonstrate the effectiveness of BPNet on three challenging NLVL benchmarks (\ie, TACoS, Charades-STA, and ActivityNet Captions) by extensive ablative studies. Particularly, BPNet achieves new state-of-the-art performance over all three datasets and evaluation metrics. 


\begin{figure*}
\centering
\includegraphics[width=\textwidth]{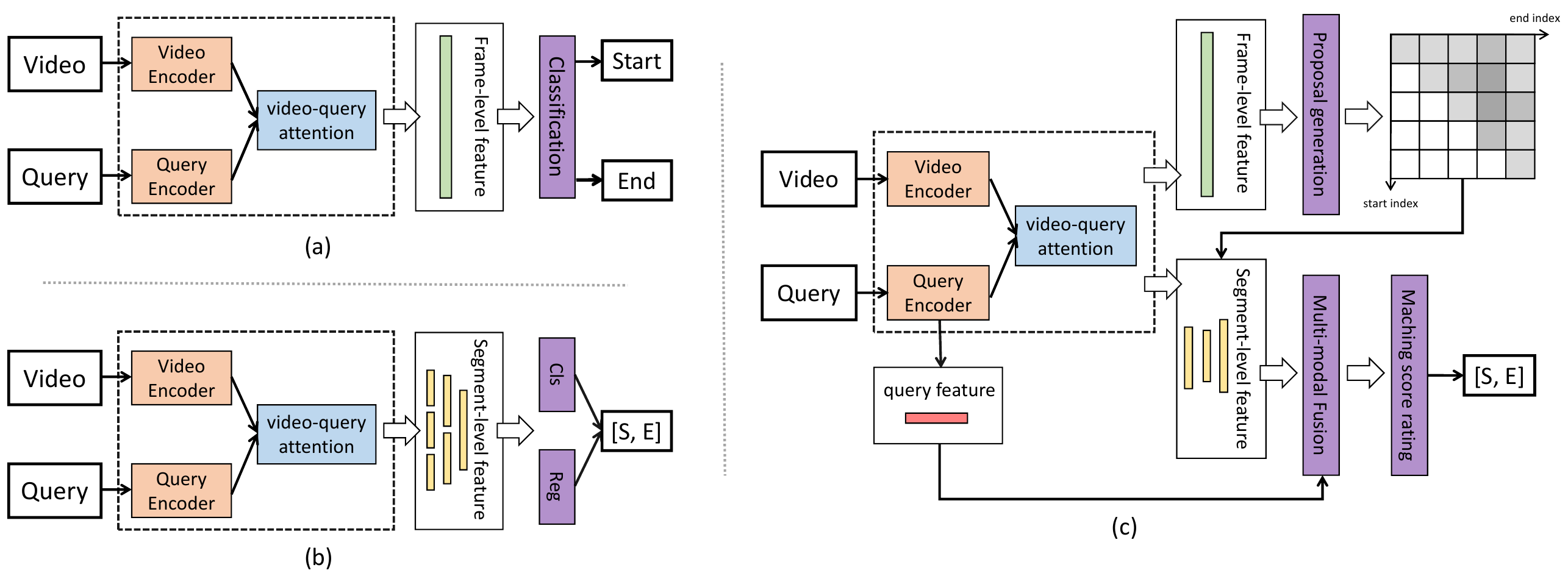}
\caption{(a) A standard framework of anchor-free approach which conducts classification on frame-level visual feature. 
(b) A standard framework of anchor-based approach which does classification 
and regression on segment-level feature.
(c) Architecture of the proposed two-stage framework: it first utilizes an 
anchor-free method to generate segment candidates and then computes a matching score with the query, which exploits both frame-level and segment-level information.
}
\label{figure2}
\end{figure*}

\section{Related Work}
\textbf{Natural Language Video Localization.}
The task of natural language video localization (NLVL) aims at predicting the start and end time of the video moment depicted by a language query within the untrimmed video, which was introduced in~\cite{DBLP:conf/iccv/HendricksWSSDR17,DBLP:conf/iccv/GaoSYN17}.
Current existing methods can be roughly grouped into two categories according to how the video segments are detected, namely \emph{anchor-based} methods and \emph{anchor-free} methods.

The anchor-based approaches~\cite{DBLP:conf/iccv/GaoSYN17,DBLP:conf/iccv/HendricksWSSDR17,DBLP:conf/emnlp/HendricksWSSDR18,DBLP:conf/mm/LiuWN0CC18,DBLP:conf/sigir/LiuWN0CC18,DBLP:conf/aaai/Xu0PSSS19,DBLP:conf/cvpr/ZhangDWWD19} solve the NLVL task by matching the predefined video moment proposals (\eg, in sliding window manner) with the language query and choose the best matching video segment as the final result.
\citet{DBLP:conf/iccv/GaoSYN17} proposed a Cross-modal
Temporal Regression Localizer (CTRL) model. It takes video moments predefined through sliding windows as input and jointly models text
query and video clips, then outputs alignment scores and action
boundary regression results for candidate clips. 
\citet{DBLP:conf/iccv/HendricksWSSDR17} proposed the Moment Context Network (MCN) which effectively localizes natural language queries in videos by integrating local and global video features over time.
To improve the performance of the anchor-based method, some works devote to improve the quality of the proposals.
\citet{DBLP:conf/aaai/Xu0PSSS19} injected text features early on when generating clip proposals to eliminate unlikely clips and thus speed up processing and boost performance.
\citet{DBLP:conf/cvpr/ZhangDWWD19} proposed to explicitly model moment-wise temporal relations as a structured graph and devised
an iterative graph adjustment network to jointly learn the
best structure in an end-to-end manner.
The others mainly worked on designing a more effective multi-modal interaction network. 
\citet{DBLP:conf/mm/LiuWN0CC18} utilized a language-temporal
attention network to learn the word attention based on
the temporal context information in the video.
\citet{DBLP:conf/sigir/LiuWN0CC18} designed a memory attention model to dynamically compute the visual attention over the query and its context
information. However, these models are sensitive to the heuristic rules (\eg, the number and the size of anchors) and suffer from inefficiency because of the dense sampling video segment candidates.

The anchor-free approaches~\cite{DBLP:conf/aaai/YuanM019,LuCTLX19,DBLP:conf/aaai/ChenLTXZTL20,ChenCMJC18,DBLP:conf/acl/ZhangSJZ20} directly predict the probabilities for each frame whether the
frame is the boundary frame of the ground-truth video segment. 
Without pre-defined size-fixed candidates, anchor-free approaches are flexible to adapt to the videos with variant length. 
\citet{DBLP:conf/aaai/YuanM019} directly regressed the temporal
coordinates from the global attention outputs. 
\citet{DBLP:conf/acl/ZhangSJZ20} regarded the NLVL task as a span-based QA
problem by treating the input video as a text passage and directly regressed 
the start and end points.
In order to further improve the performance, some works focus on eliminating the problem of imbalance of the positive and negative samples. 
\citet{LuCTLX19} and \citet{DBLP:conf/aaai/ChenLTXZTL20} regarded all frames falling in the ground
truth segment as foreground, and each foreground frame regresses the unique distances from its location to bi-directional ground truth boundaries.
Our BPNet focuses on the other weakness of the anchor-free approach that it is hard 
to model the segment-level multi-modal features. BPNet takes  
both frame-level and segment-level visual information into consideration to further improve the performance.

There are also some other works~\cite{DBLP:conf/aaai/HeZHLLW19,DBLP:conf/cvpr/WangHW19} solving the NLVL task by reinforcement learning, which formulates the selection of start and end time points as a sequential decision making process.
\textbf{Anchor-based and Anchor-free Object Detection.}
The development of NLVL is inspired by the success of object detection approaches. 
Object detection aims to obtain a tight bounding box and a class label 
for each object.
It can be categorized into anchor-based and anchor-free approaches according to 
the way to localize an object. 
Traditional anchor-based models~\cite{DBLP:conf/nips/RenHGS15,DBLP:conf/nips/DaiLHS16} 
have dominated this area for many years, 
which place a series of anchors (bounding boxes) uniformly and do the classification and regression to determine the position and class for the objects.
Recently, researches on anchor-free models~\cite{DBLP:conf/eccv/LawD18,DBLP:conf/iccv/DuanBXQH019}
are becoming prosperous, which have 
been promoted by the development of keypoint detection. 
The anchor-free methods directly predict the keypoints and group them together 
to determine the object. 
By the comparison of the two approaches, anchor-free methods are more flexible to locate objects with arbitrary geometry but have lower performance in contrast to the anchor-based methods because of the misalignment of keypoints. 
\citet{DBLP:conf/eccv/Duan} presented an anchor-free two-stage object detection framework termed CPN that extracts the keypoints and composes them into object proposals, then two-step classification is used to filter out the false positives.
The BPNet borrows the similar idea from CPN which inherits the merits of both anchor-free and anchor-based approaches.

\section{Proposed Approach}
We define the NLVL task as follows. Given an untrimmed video as $V=\{f_t\}^T_{t=1}$ and a language query 
as $Q=\{w_n\}^M_{m=1}$, where $T$ and $M$ are the number of video frames and query words, NLVL needs to predict the 
start time $t_s$ and the end time $t_e$ of the video segment described by the language query $Q$.  
For each video, we extract its visual features $\bm{V} = \{\bm{v}_t\}^T_{t=1}$ by a pre-trained 3D ConvNet~\cite{DBLP:conf/iccv/TranBFTP15}. For each query, we initialize the word features $\bm{Q} = \{\bm{w}_n\}^M_{m=1}$ using the GloVe embeddings~\cite{DBLP:conf/emnlp/PenningtonSM14}. 

As shown in Figure~\ref{figure2}(c), our framework operates in two steps: boundary proposal generation and visual-language matching. Specifically, in the first step, it uses an anchor-free method to extract video segment proposals. In the second step, it fuses each segment proposal with the language and computes a matching score. The proposal with the highest matching score will be chosen as the correct segment. 

In this section, we introduce the architecture of our BPNet. In Section~\ref{BPG}, we first describe the boundary proposal generation phase. In Section~\ref{VLM}, we then present the visual-language matching phase. Finally, in Section~\ref{training}, we show the training and inference processes of BPNet in detail.

\subsection{Boundary Proposal Generation}\label{BPG}
The first stage is an anchor-free proposal extraction process, in which we generate a series of video segment proposals. Different from 
the existing anchor-based approaches, our generated proposals are of 
high quality because we utilize an effective anchor-free approach. 
We follow the anchor-free backbone of Video Span Localization Network (VSLNet)~\cite{DBLP:conf/acl/ZhangSJZ20}, which addressed the NLVL task as a span-based QA task. It is worth noting that our proposed BPNet is 
a universal framework that can incorporate any other anchor-free 
approach.

\begin{figure}[t]
    \centering
    \includegraphics[width=0.48\textwidth]{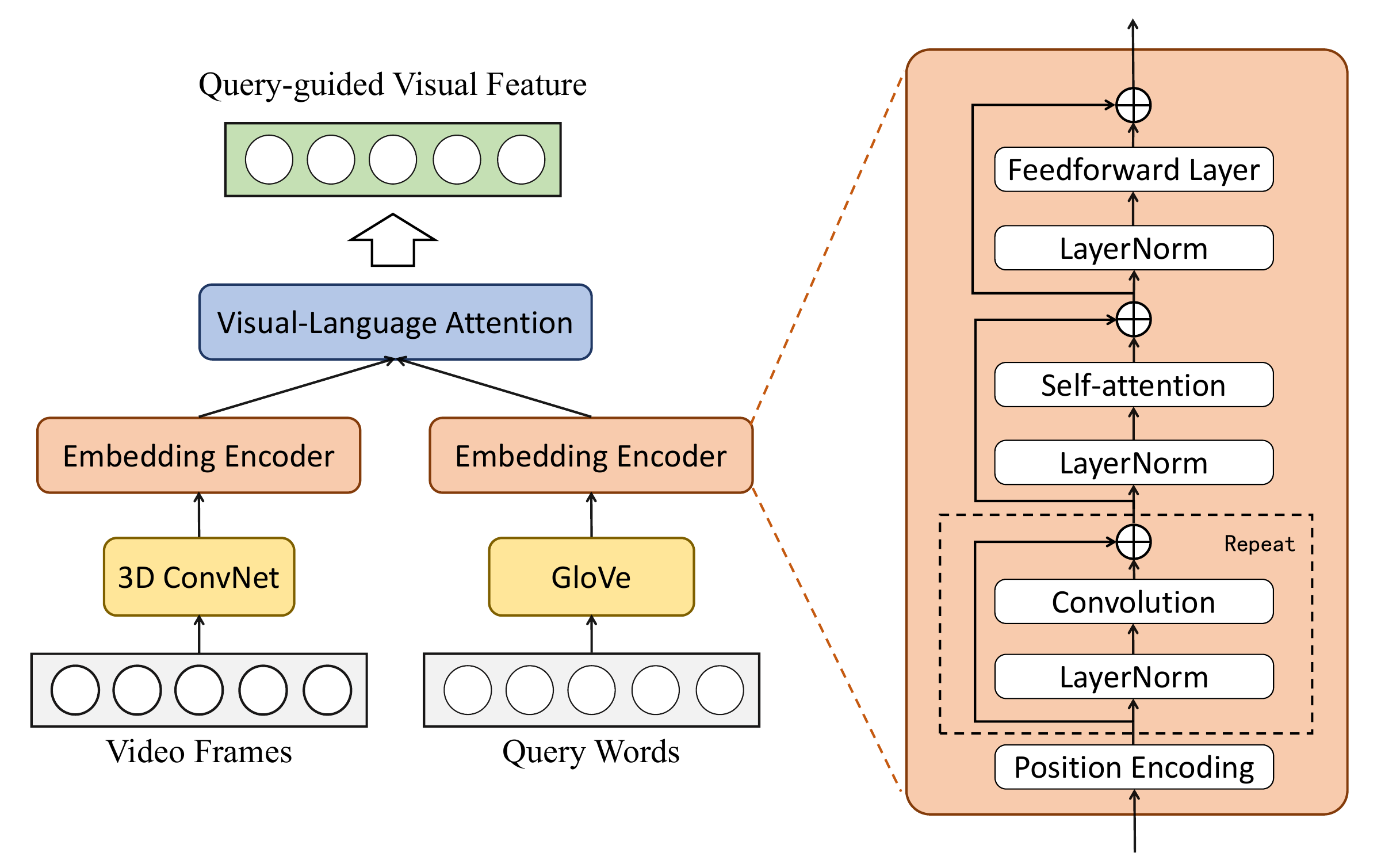} 
    \caption{The backbone of BPNet (a variant of the QANet). 
    The embedding encoder consists of several conv-layers, a self-attention layer and a feed-forward layer. For each layer, layer normalization and residual connection are employed.
    }
    \label{figure3}
\end{figure}

The first stage of BPNet consists of three components: 

\noindent\textbf{Embedding Encoder Layer.}
We use a similar encoder in QANet~\cite{DBLP:conf/iclr/YuDLZ00L18}. 
The input of this layer is visual features $\bm{V} \in \mathbb{R}^{T\times d_{v}}$ and text query feature $\bm{Q}\in\mathbb{R}^{M\times d_{q}}$. 
We project them into the same dimension and feed them into the embedding encoder layer to integrate contextual information.
\begin{equation} \label{equation1}
\begin{split}
\bm{V}^{'} & =\text{EmbeddingEncoder}(\bm{V}\bm{W}_{v}), \\
\bm{Q}^{'} & =\text{EmbeddingEncoder}(\bm{Q}\bm{W}_{q}),
\end{split}
\end{equation}
where $\bm{W}_{v}\in\mathbb{R}^{d_{v}\times d}$, $\bm{W}_{q}\in\mathbb{R}^{d_{q}\times d}$ are project matrices. 
Notice that the biases for transformation layers are omitted for clarity (the same below). 
As shown in Figure~\ref{figure3}, the embedding encoder layer consists of multiple components, 
including four convolution layers, multi-head attention layer, layer normalization layer and feed-forward layer with the residual connection.
The output of the embedding encoder layer $\bm{V}^{'} \in \mathbb{R}^{T\times d}$ and $\bm{Q}^{'}\in\mathbb{R}^{M\times d}$ are refined visual and language features that encode the interaction inside each modality.

\noindent\textbf{Visual-Language Attention Layer.}
This layer calculates vision-to-language attention and language-to-vision attention weights and encodes the two modal features together. 
Specifically, if first computes a similarity matrix $\mathcal{S}\in\mathbb{R}^{T\times M}$, where the element $\mathcal{S}_{ij}$ 
indicates the similarity between the frame $f_{i}$ and the word $w_{j}$. 
Then the two attention weights $\bm{A}$ and $\bm{B}$ are computed:
\begin{equation}
\bm{A} = \mathcal{S}_{row}\cdot\bm{Q}^{'}, \quad \bm{B} = \mathcal{S}_{row}\cdot\mathcal{S}_{col}^{T}\cdot\bm{V}^{'},
\label{equation2}
\end{equation}
where $\mathcal{S}_{row}$ and $\mathcal{S}_{col}$ are the row and column-wise normalization of $\mathcal{S}$.
We then model the interaction between the video and the query by the cross-modal attention layer:
\begin{equation}
\bm{V}^{q}=\text{FFN}([\bm{V}^{'};\bm{A};\bm{V}^{'}\odot\bm{A};\bm{V}^{'}\odot\bm{B}]),
\label{equation3}
\end{equation}
where $\odot$ is the element-wise multiplication, and $[\cdot]$ is the concatenation operation. The $\text{FFN}$ represents feed-forward layer. The output of this layer $\bm{V}^{q}$ encodes the visual feature with query-guided attention.

\noindent\textbf{Proposal Generation Layer.}
After getting the query-guided visual feature $\bm{V}^{q}$, 
we now generate proposals by using two stacked LSTMs, the hidden states of which are fed into two feed-forward layers to compute the start and end scores:
\begin{alignat}{2}
\bm{H}^{s}&= \text{LSTM}_{s}(\bm{V}^{q}), \quad &\bm{S}^{s}&= \text{FFN}(\bm{H}^{s}), \\
\bm{H}^{e}&= \text{LSTM}_{e}(\bm{H}^{s}), \quad &\bm{S}^{e}&= \text{FFN}(\bm{H}^{e}),
\label{equation45}
\end{alignat}
where $\bm{H}^{s}$ and $\bm{H}^{e}$ are the hidden states of the 
$\text{LSTM}_{s}$ and $\text{LSTM}_{e}$; $\bm{S}^{s}$ and $\bm{S}^{e}$ denote the logits of start and end boundaries computed by a feed-forward layer.

Then, we compute the joint probability of start and end points using 
matrix multiplication:
\begin{equation}
\begin{split}
\bm{P}_{s}&=\text{softmax}(\bm{S}^{s}), \\
\bm{P}_{e}&=\text{softmax}(\bm{S}^{e}), \\
\bm{M}^{p} &= \bm{P}_{s}^{T}\bm{P}_{e},
\end{split}
\label{equation6}
\end{equation}
where $\bm{P}_{s}$ and $\bm{P}_{e}$ are probability distributions of the start and end boundaries.
$\bm{M}^{p}$ is a two-dimensional score map whose element indicates the predicted probability 
of each video segment, \eg, $\bm{M}^{p}_{ij}$ denotes the score for the segment from start boundary $i$ to 
end boundary $j$. 
We sample the $N$ highest position on the score map $\bm{M}^{p}$ and treat the corresponding segments as 
candidates.

\subsection{Visual-Language Matching}\label{VLM}
\textbf{Visual-Language Fusion Layer.}
Given $N$ video segment candidates for video $V$, we capture the candidate features
$\bm{C}$ from the visual feature $\bm{V}^{'}$ in Eq.~\eqref{equation1}.
The generated video segment candidates have different lengths in the temporal dimension,  
hence we transform the candidate features into identical length using the temporal weighted pooling. 
We also obtain sentence-level query feature by weighted pooling over the word-level features. 
Then, we fuse them by concatenation followed by a feed-forward layer.
\begin{equation}
\begin{split}
\bm{\widetilde{C}}_{i}&= \text{AvgPooling}(\bm{W}_{c}\bm{C}_{i}), \\
\bm{\widetilde{Q}}&= \text{AvgPooling}(\bm{W}_{q}\bm{Q}^{'}), \\
\bm{F}&= \text{FFN}([\bm{\widetilde{C}}_{i},\bm{\widetilde{Q}}]),
\end{split}
\label{equation7}
\end{equation}
where $\bm{W}_{c}$ and $\bm{W}_{q}$ are learnable weights and 
$[\cdot]$ is the concatenation operation.
This layer is to encode the segment-level information of video and fuse  
the visual and language feature for the subsequent process.

\noindent\textbf{Matching Score Rating Layer.}
Taking the multi-modal feature as input, this layer predicts the matching score for each video segment proposal and the language query. 
The most matched proposal will be chosen as the final result. 
This layer consists of two feed-forward layers followed by ReLU and sigmoid activation respectively: 
\begin{equation}
\hat{s}_{i} = \text{sigmoid}(\bm{W}_{2} \text{ReLU} (\bm{W}_{1}\bm{F})),
\label{equation8}
\end{equation}
where $\hat{s}_{i}$ indicates the predicted matching score of the $i$-th candidates. 
We argue that the matching scores are positively associated with the temporal
IoU scores between candidates and the ground-truth video segment. 
Therefore, we use the IoU scores as the ground-truth labels to supervise the training process. 
As a result, 
the matching score rating problem turns into an IoU regression problem. 

\subsection{Training and Inference}\label{training}
\textbf{Training.} Each training sample consists of an untrimmed video, a language query and the ground-truth 
video segment. 
Specifically, for each video frame with the frame-level feature, two class labels indicated whether or not the 
frame is the start or the end boundary are assigned.
For each segment candidate with the segment-level feature, we compute the temporal IoU between the 
candidate and the ground-truth segment as the matching score.

There are two loss functions for the boundary proposal generation stage and visual-language matching stage:

\noindent\textbf{Boundary Classification Loss:}
\begin{equation}
\mathcal{L}_{cls} = f_{CE}(P_{s},Y_{s}) + f_{CE}(P_{e},Y_{e}),
\label{equation9}
\end{equation}
where the $f_{CE}$ is a binary cross entropy loss function. 
$Y_{s}$ and $Y_{e}$ are ground-truth labels for the start and end boundaries. 

\noindent\textbf{Matching Regression Loss:}
\begin{equation}
\mathcal{L}_{reg} = f_{MSE}(\hat{s},s_{IoU}),
\label{equation10}
\end{equation}
where $f_{MSE}$ is a $L2$ loss function and $s_{IoU}$ is the ground-truth temporal IoU scores. 

Thus, the final loss is a multi-task loss combining the $\mathcal{L}_{cls}$ and $\mathcal{L}_{reg}$, \ie,
\begin{equation}
\mathcal{L} = \mathcal{L}_{cls} + \lambda \times \mathcal{L}_{reg},
\label{equation11}
\end{equation}
where $\lambda$ is a hyper-parameter that balances the two losses. 

\noindent\textbf{Inference.} Given a video and a language query, 
we forward them through the network and obtain $N$ segment candidates 
with their corresponding matching scores $\hat{s}$. Then, we rank the 
$\hat{s}$ and select the candidate with the highest score as the final 
result.

\section{Experiments}
\subsection{Datasets}\label{Dataset}
We evaluate our BPNet on three public benchmark datasets: 1) \textbf{Charades-STA}~\cite{DBLP:conf/iccv/GaoSYN17}: It is built on Charades and contains 6,672 videos of daily indoors activities. Charades-STA contains 16,128 sentence-moment pairs in total, where 12,408 pairs are for training and 3,720 pairs for testing. The average duration of the videos is 30.59s and the average duration of the video segments is 8.22s. 2) \textbf{TACoS}: It consists of 127 videos of cooking activities.
For video grounding task, it contains 18818 sentence-moment pairs in total. 
Followed by the split setting in~\cite{DBLP:conf/iccv/GaoSYN17}, we use 10,146, 4,589, 4,083 
for training, validation and testing respectively.
The duration of the videos is 287.14s on average and the average length of the video segments 
is 5.45s. 3) \textbf{ActivityNet Captions}~\cite{DBLP:conf/iccv/KrishnaHRFN17}: It contains around 20k open domain videos for video grounding task. \
We follow the split in~\cite{DBLP:conf/aaai/YuanM019}, which consists of 37,421 
sentence-moment pairs for training and 17,505 for testing.
The average duration of the videos is 117.61s and the average length of the video segments 
is 36.18s.

\subsection{Evaluation Metrics}\label{Metrics}
Following the prior works, we adopt ``R@$n$, IoU=$\theta$" and ``mIoU" as evaluation metrics. 
Specifically,``R@$n$, IoU=$\theta$" represents the percentage of the testing samples that 
have at least one of the top-N results whose IoU with the ground-truth is larger than $\theta$. 
The ``mIoU" means the average IoU with ground truth over all testing samples. 
In all the experiments, we set $n$ = 1 and $\theta\in \{0.3, 0.5, 0.7\}$. 

\subsection{Implementation}\label{Implementation}
We down-sample frames for each video and extract visual features 
using C3D~\cite{DBLP:conf/iccv/TranBFTP15} network pretrained on Sports-1M. Then we reduce the features 
to 500 dimension by PCA. 
For language query, we initialize each word with 300d GloVe vectors and all word embeddings are fixed during training.
The dimension of the intermediate layer in BPNet is set to 128. 
The number of convolution blocks in embedding encoder is 4 and the kernel size is set to 7. 
The number of boundary proposals is 128 for training and 8 for testing. 
For all datasets, we trained the model for 100 epochs with 
batch size of 32. 
Dropout and early stopping strategies are adopted to prevent overfitting. 
We implement our BPNet on Tensorflow. 
The whole framework is trained by Adam optimizer with learning rate 0.0001. 

\subsection{Comparisons with the State-of-the-Arts}\label{Comparison}
\noindent\textbf{Settings.} We compare the proposed BPNet with several state-of-the-art NLVL methods on three datasets. These methods are grouped into three categories by the viewpoints of anchor-based and anchor-free approach: 
 1) Anchor-based models: 
\textbf{VSA\_RNN}, \textbf{VSA\_STV}, \textbf{CTRL}~\cite{DBLP:conf/iccv/GaoSYN17}, \textbf{ACRN}~\cite{DBLP:conf/sigir/LiuWN0CC18}, \textbf{ROLE}~\cite{DBLP:conf/mm/LiuWN0CC18}, \textbf{MCF}~\cite{DBLP:conf/ijcai/WuH18}, \textbf{ACL}~\cite{DBLP:conf/wacv/GeGCN19}, \textbf{SAP}~\cite{DBLP:conf/aaai/ChenJ19a}, 
\textbf{QSPN}~\cite{DBLP:conf/aaai/Xu0PSSS19}, 
\textbf{TGN}~\cite{ChenCMJC18},
\textbf{MAN}~\cite{DBLP:conf/cvpr/ZhangDWWD19}. 
2) Anchor-free models:
\textbf{L-Net}~\cite{DBLP:conf/aaai/Chen0CJL19}, \textbf{ABLR-af}, \textbf{ABLR-aw}~\cite{DBLP:conf/aaai/YuanM019}, \textbf{DEBUG}~\cite{LuCTLX19},
\textbf{ExCL}~\cite{DBLP:conf/naacl/GhoshAPH19},
\textbf{GDP}~\cite{DBLP:conf/aaai/ChenLTXZTL20}, \textbf{VSLNet}~\cite{DBLP:conf/acl/ZhangSJZ20}. 
3) Others:
\textbf{RWM}~\cite{DBLP:conf/aaai/HeZHLLW19}, \textbf{SM-RL}~\cite{DBLP:conf/cvpr/WangHW19}.

The results on three benchmarks are reported in Table~\ref{table1} to Table~\ref{table3}. We can observe that our BPNet achieves new state-of-the-art performance over all metrics and benchmarks.
Table~\ref{table1} summarizes the results on Charades-STA. 
We can observe that BPNet outperforms all the baselines 
in all metrics.
Specifically, we observe that BPNet works well in even stricter metrics, 
\eg, BPNet achieved a significant 2.59 absolute improvement in IoU@0.7 
compared to the second result, which demonstrates the effectiveness of 
our model.
For a fair comparison with VSLNet~\cite{DBLP:conf/acl/ZhangSJZ20}, we use both C3D and I3D~\cite{DBLP:conf/cvpr/CarreiraZ17} visual features. 
VSLNet is a typical anchor-free model with state-of-the-art performance 
whose architecture is roughly described in Figure~\ref{figure2} (a). 
Specifically, we implement the VSLNet with C3D feature followed the settings they reported. 
VSLNet extracts the frame-level feature using the backbone of QANet and utilizes two 
LSTM to classify the start/end boundary. 
Our model outperforms VSLNet in all metrics on Charades-STA. It is mainly because that 
BPNet better models the segment-level visual information between the boundaries.

\begin{table}[t]
    \centering
    \setlength{\tabcolsep}{1.1mm}{
    \begin{tabular}{l|c|c|c|c|c}
    \hline
    Methods & Feature  & IoU=0.3 & IoU=0.5 & IoU=0.7 & mIoU  \\ \hline
    VSA-RNN & C3D      & --      & 10.50    & 4.32    & --    \\
    VSA-STV & C3D      & --      & 16.91   & 5.81    & --    \\
    CTRL    & C3D      & --      & 23.63   & 8.89    & --    \\
    ROLE    & C3D      & 25.26   & 12.12   & --      & --    \\
    ACL-K     & C3D      & --   & 30.48   & 12.20   & --    \\
    SAP     & C3D      & --      & 27.42   & 13.36   & --    \\
    RWM     & C3D      & --      & 36.70    & --      & --    \\
    SM-RL   & C3D      & --      & 24.36   & 11.17   & --    \\
    QSPN    & C3D      & 54.70    & 35.60    & 15.80    & --    \\
    DEBUG   & C3D      & 54.95   & 37.39   & 17.92   & 36.34    \\
    GDP     & C3D      & 54.54   & 39.47   & 18.49   & --     \\
    VSLNet  & C3D      & 54.38   & 28.71   & 15.11   & 37.07   \\
    BPNet    & C3D      & \textbf{55.46}   & \textbf{38.25}   & \textbf{20.51}   & \textbf{38.03}    \\ \hline
    ExCL    & I3D      & --      & 44.10    & 22.40    & --    \\
    MAN     & I3D      & --        & 46.53   & 22.72    & --       \\
    VSLNet  & I3D      & 64.30    & 47.31   & 30.19   & 45.15 \\
    BPNet    & I3D      & \textbf{65.48}  & \textbf{50.75}   & \textbf{31.64}   & \textbf{46.34}\\ \hline
    \end{tabular}}
    \caption{Performance (\%) of ``R@$n$, IoU=$\theta$" and ``mIoU"  compared with the state-of-the-art NLVL models on Charades-STA.} 
    \label{table1}
\end{table}

\begin{table}[!h]
    \centering
     \setlength{\tabcolsep}{1.5mm}{
        \begin{tabular}{l|c|c|c|c}
        \hline
        Methods  & IoU=0.3 & IoU=0.5 & IoU=0.7 & mIoU  \\ \hline
        VSA-RNN  & 6.91    & --      & --      & --    \\
        VSA-STV  & 10.77   & --      & --      & --    \\
        CTRL     & 18.32   & 13.30     & --      & --    \\
        ACRN     & 19.52   & 14.62      & --      & --    \\
        MCF      & 18.64   & --      & --      & --    \\
        SM-RL    & 20.25   & 15.95      & --      & --    \\
        ACL      & 22.07   & 17.78      & --      & --    \\
        SAP      & --      & 18.24      & --      & --    \\
        L-NET    & --      & --      & --      & 13.41 \\
        TGN      & 21.77    & 18.90   &  --      & -- \\
        ABLR-aw   & 18.90    & 9.30      & --      & 12.50  \\
        ABLR-af   & 19.50    & --      & --      & 13.40  \\
        DEBUG    & 23.45   & 11.72      & --      & 16.03 \\
        GDP    & 24.14   & --          & --      & 16.18 \\
        VSLNet   & 22.88   & 18.98   & 14.01   & 18.01 \\
        BPNet     & \textbf{25.96}   & \textbf{20.96}   & \textbf{14.08}   & \textbf{19.53} \\ \hline
        \end{tabular}}
        \caption{Performance (\%) of ``R@$n$, IoU=$\theta$" and ``mIoU" compared with the state-of-the-art NLVL models on TACoS.} 
        \label{table2}
\end{table}
  
\begin{table}[!h]
    \centering
    \setlength{\tabcolsep}{1.45mm}{
    \begin{tabular}{l|c|c|c|c}
    \hline
    Methods  & IoU=0.3 & IoU=0.5 & IoU=0.7 & mIoU  \\ \hline
    TGN         & 43.81   & 27.93   & --      & --    \\
    QSPN        & 45.30    & 27.70    & 13.60      & --    \\
    RWM         & --      & 36.90    & --      & --    \\
    ABLR-af     & 53.65   & 34.91   & --      & 35.72 \\
    ABLR-aw     & 55.67   & 36.79   & --      & 36.99 \\
    DEBUG       & 55.91   & 39.72   & --      & 39.51 \\
    GDP         & 56.17   & 39.27   & --       & 18.49 \\
    VSLNet      & 55.17   & 38.34   & 23.52   & 40.53 \\
    BPNet        & \textbf{58.98}   & \textbf{42.07}   & \textbf{24.69}   & \textbf{42.11} \\ \hline
    \end{tabular}}
    \caption{Performance (\%) of ``R@$n$, IoU=$\theta$" and ``mIoU" compared with the state-of-the-art NLVL models on ActivityNet Captions.} 
    \label{table3}
\end{table}

\begin{table}[t]
    \centering
    \setlength{\tabcolsep}{0.6mm}{
    \begin{tabular}{l|c|c|c|c}
    \hline
    \multicolumn{5}{c}{Charades-STA} \\
    \hline
    Methods  & IoU=0.3 & IoU=0.5 & IoU=0.7 & mIoU                      \\ \hline
    VSLNet (anchor-free)   & 54.38   &28.71    &15.11      &37.07                    \\
    Our (anchor-based)     &55.97    &34.81   & 15.46      & 35.94     \\
    BPNet     &  \textbf{55.46}  & \textbf{38.26}   &  \textbf{20.51}     & \textbf{38.03}   \\ \hline
    \multicolumn{5}{c}{TACoS} \\
    \hline
    Methods  & IoU=0.3 & IoU=0.5 & IoU=0.7 & mIoU                      \\ \hline
    VSLNet (anchor-free)  & 22.88   & 18.98   & 14.01   & 18.01                     \\
    Our (anchor-based)     &22.39    &14.67    & 7.42      & 15.63      \\
    BPNet    & \textbf{25.96}   & \textbf{20.96}   & \textbf{14.08}   & \textbf{19.53}        \\ \hline
    \multicolumn{5}{c}{ActivityNet Captions} \\
    \hline
    Methods  & IoU=0.3 & IoU=0.5 & IoU=0.7 & mIoU                       \\ \hline
    VSLNet (anchor-free)    & 55.17   & 38.34   & 23.52   & 40.53                     \\
    Our (anchor-based)     &58.75    &40.52    & 18.74      & 39.66         \\
    BPNet     & \textbf{58.98}  & \textbf{42.07}  & \textbf{24.69}   &\textbf{42.11}         \\ \hline
    \end{tabular}}
    \caption{Performance (\%) comparisons of the anchor-free model, anchor-based model and BPNet with the same backbone on three benchmarks. } 
    \label{table4}
\end{table}

\begin{figure*}
    \centering
    \includegraphics[height=0.46\textheight,width=\textwidth]{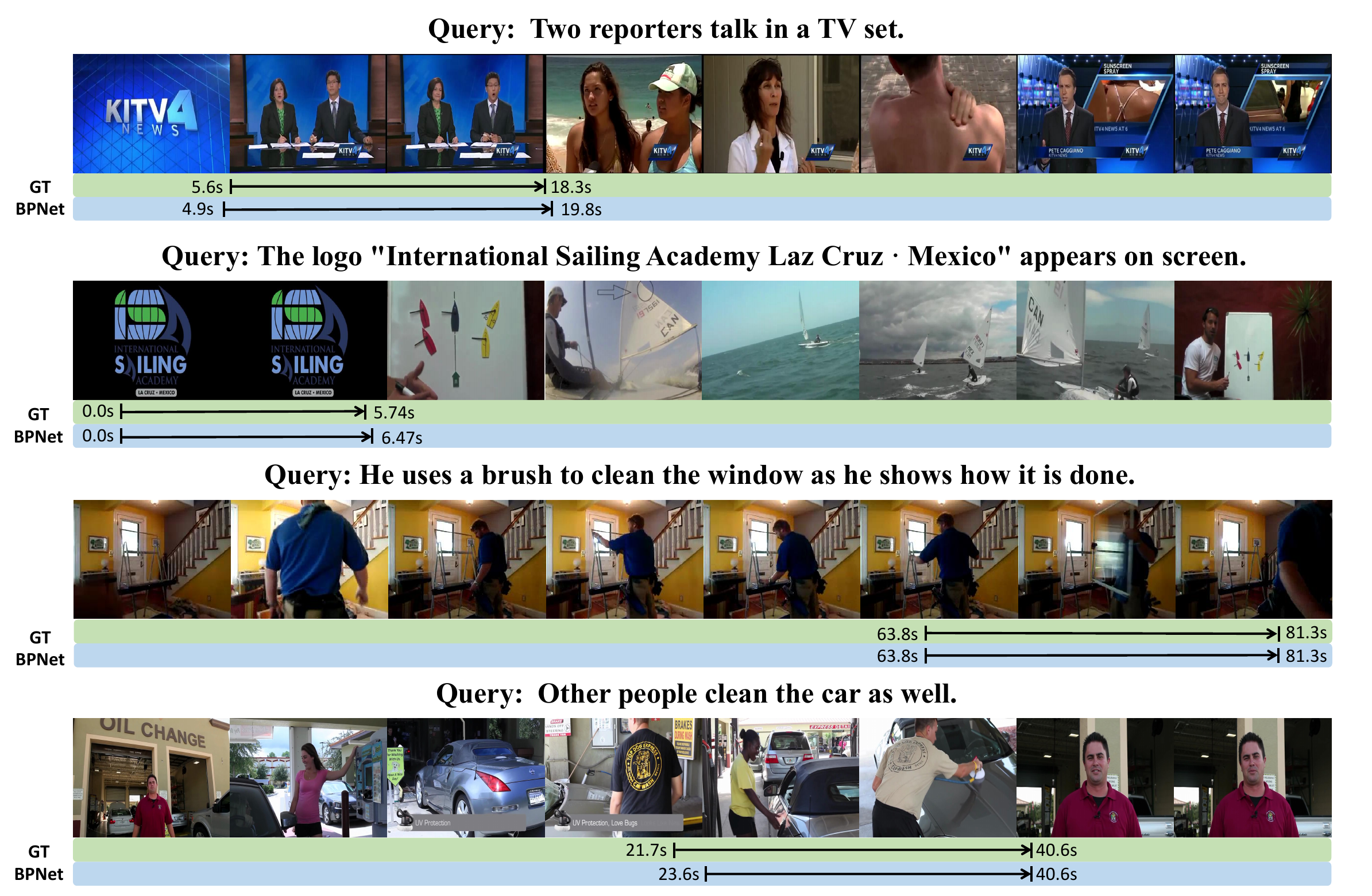}
    \caption{The qualitative results of BPNet on ActivityNet Captions.}
    \label{figure4}
\end{figure*}

The results on TACoS and ActivityNet Captions are summarized in Table~\ref{table2} and Table~\ref{table3}. 
Note that the videos in TACoS have a longer average duration and the ground-truth video segments in ActivityNet Captions have a longer average length. 
BPNet significantly outperforms the other methods on both benchmarks with the 
C3D feature, which demonstrates that BPNet is highly adaptive to videos and 
segments with diverse lengths. 
The qualitative results of BPNet is illustrated in Figure~\ref{figure4}.

It is worth noting that BPNet can utilize a more effective anchor-free backbone 
such as~\cite{DBLP:conf/aaai/ZhangPFL20,DBLP:conf/cvpr/ZengXHCTG20} to further improve the performance. 
Even so, we take into account the simplicity and efficiency and choose the VSLNet as the backbone. 

\begin{table}[htbp]
\centering
\setlength{\tabcolsep}{1.6mm}{
\begin{tabular}{c|ccccc|c|c}
\hline
\multirow{2}{*}{Methods} &
  \multicolumn{5}{c|}{Length of Windows} &
  \multirow{2}{*}{num} &
  \multirow{2}{*}{mIoU} \\ \cline{2-6}
 &
  \multicolumn{1}{c|}{128} &
  \multicolumn{1}{c|}{64} &
  \multicolumn{1}{c|}{32} &
  \multicolumn{1}{c|}{16} &
  8 &
   &
   \\ \hline
\multirow{6}{*}{\begin{tabular}[c]{@{}c@{}}Anchor-based \\ Model\end{tabular}} &
  \multicolumn{1}{c|}{\checkmark} &
  \multicolumn{1}{c|}{} &
  \multicolumn{1}{c|}{} &
  \multicolumn{1}{c|}{} &
   &
  5 &
  3.92 \\
 &
  \multicolumn{1}{c|}{} &
  \multicolumn{1}{c|}{\checkmark} &
  \multicolumn{1}{c|}{} &
  \multicolumn{1}{c|}{} &
   &
  13 &
  5.88 \\
 &
  \multicolumn{1}{c|}{} &
  \multicolumn{1}{c|}{} &
  \multicolumn{1}{c|}{\checkmark} &
  \multicolumn{1}{c|}{} &
   &
  29 &
  8.62 \\
 &
  \multicolumn{1}{c|}{} &
  \multicolumn{1}{c|}{} &
  \multicolumn{1}{c|}{} &
  \multicolumn{1}{c|}{\checkmark} &
   &
  61 &
  12.32 \\
 &
  \multicolumn{1}{c|}{} &
  \multicolumn{1}{c|}{} &
  \multicolumn{1}{c|}{} &
  \multicolumn{1}{c|}{} &
  \checkmark &
  125 &
  13.50 \\
 &
  \multicolumn{1}{c|}{\checkmark} &
  \multicolumn{1}{c|}{\checkmark} &
  \multicolumn{1}{c|}{\checkmark} &
  \multicolumn{1}{c|}{\checkmark} &
  \checkmark &
  233 &
  15.65 \\ \hline
BPNet &
  \multicolumn{5}{c|}{no limit} &
  8 &
  \textbf{19.53} \\ \hline
\end{tabular}}
\caption{Performance (\%) comparisons with the anchor-based model in different settings on TACoS.}
\label{table5}
\end{table}

\begin{table}[h]
    \centering
    \setlength{\tabcolsep}{2mm}{
    \begin{tabular}{c|c|c|c|c}
    \hline
    Method  & IoU=0.3 & IoU=0.5 & IoU=0.7 & mIoU  \\ \hline
    BPNet w/o vlf &  55.05       &  34.78       & 18.84        & 37.92      \\
    BPNet & \textbf{55.46}   & \textbf{38.25}   & \textbf{20.51}   & \textbf{38.03} \\ \hline
    \end{tabular}}
    \caption{Performance (\%) comparisons on Charades-STA. BPNet w/o vlf represent the BPNet without 
    the Visual-Language Fusion layer.}
    \label{table6}
\end{table}
\subsection{Ablation Study}
\label{ablation}
In this section, we conduct ablative experiments with different 
variants to better investigate our approach. 

\noindent\textbf{Anchor-based vs. Anchor-free.}
To evaluate the effectiveness of our two-stage model, 
we compare BPNet with both anchor-free and anchor-based models. 
For a fair comparison, we designed an anchor-based model with the 
same backbone as BPNet. Specifically, we sampled a series of 
bounding boxes over the temporal dimension and conduct the following 
matching process. Since VSLNet has the same backbone as 
BPNet, we use it as the anchor-free setting.

\noindent\emph{Results.}
The results of the three models are reported in Table~\ref{table4}. 
We can observe that our BPNet outperforms all the baselines with 
the same backbone. 
In particular, we observe that our implementation of the anchor-based 
model works well on looser metrics (\eg, anchor-based)
while the anchor-free model does 
better on strict metrics. We think that's because the anchor-based model has wider coverage bounding boxes. 
BPNet performed better than both models over all metrics.

\noindent\textbf{Quality of the Candidates.}
We compare the quality of the candidates generated by our BPNet and 
by dense sampling bounding boxes. 
The results conducted on TACoS are reported in Table~\ref{table5}. 
In the anchor-based setting, we perform bounding boxes uniformly on video frames. 
The lengths of sliding windows are 8, 16, 32, 64 and 128, window's overlap is 0.75. We evaluate the anchor-based model with different window lengths, which results 
in variant number of candidates. 
We notice that our BPNet only generated 8 candidates which are much less than the 
anchor-based setting but achieved a higher performance. 
This indicates that BPNet can generate high-quality candidates. 

\noindent\textbf{With vs. Without Visual-Language Fusion Layer.}
We evaluate the model without multi-modal fusion before matching score rating. 
For a fair comparison, we use the multi-modal features $V^{q}$ in Eq.~\eqref{equation3} for 
this setting. 
From Table~\ref{table6}, we can observe that the Visual-Language Fusion Layer 
improves the performance.  
The main reason is that the multi-modal fusion layer is able to 
model the segment-level video-query interaction.

\section{Conclusions}
In this paper, we propose a novel Boundary Proposal Network (BPNet) 
for natural language video localization (NLVL). 
By utilizing an anchor-free model to generate high-quality video segment 
candidates, 
we disentangle the candidate proposals from the predefined heuristic rules 
to make them adaptable to video segments with variant lengths. 
Furthermore, we jointly model the segment-level video feature and query feature, 
which further boosts the performance. 
As a result, the proposed BPNet outperforms the state-of-the-art approaches 
on three benchmark datasets. 
Moreover, BPNet is a universal framework which means that the proposal 
generation module and the visual-language matching module can be replaced by any 
other effective methods. In the future, we are going to extend this framework into other related tasks, \eg, visual grounding~\cite{chen2021ref}.

\section*{Acknowledgments} 
This work was supported by the National Key Research \& Development Project of China (2018AAA0101900), the National Natural Science Foundation of China (U19B2043, 61976185), Zhejiang Natural Science Foundation (LR19F020002, LZ17F020001), Key Research \& Development Project of Zhejiang Province(2018C03055), Major Project of Zhejiang Social Science Foundation (21XXJC01ZD), and the Fundamental Research Funds for the Central Universities.

\bibliography{my.bib}

\end{document}